\documentclass[10pt,leqno]{amsart}
\usepackage{graphicx}
\usepackage{amssymb}
\usepackage{amsmath}
\usepackage{amsfonts}
\usepackage{mathtools}
\usepackage{comment}
\usepackage{float}   
\usepackage{graphicx}
\usepackage{subcaption}
\usepackage{indentfirst,csquotes}
\usepackage{amssymb,amsthm,amsmath}
\usepackage{xcolor,paralist,hyperref,titlesec,fancyhdr,etoolbox}

\titleformat{\section}[display]{\normalfont\huge\bfseries\centering}{\centering\chaptertitlename\thechapter}{10pt}{\Large}
\titlespacing*{\section}{0pt}{0ex}{0ex}
\hypersetup{ colorlinks=true, linkcolor=black, filecolor=black, urlcolor=black }

\usepackage{lipsum}
\begin{document}
\title{ Max-Min  Neural Network Operators For Approximation of Multivariate Functions} 
\author{Abhishek Yadav$^{a}$, Uaday Singh$^{a}$, Feng Dai$^{b}$}
\date{\today}
\address{organization={$^a$Department of Mathematics, Indian Institute of Technology Roorkee},
            addressline={Roorkee, Uttrakhand, India}, 
            city={Roorkee},
            postcode={247667}, 
            state={Uttrakhand},
            country={India}\\{$^{b}$Department of Mathematical and Statistical Sciences, 
University of Alberta, 
Edmonton, AB, T6G 2G1, Canada}}
\email{abhishek703yadav@gmail.com}
\maketitle
\let\thefootnote\relax
\begin{abstract}
In this paper, we develop a multivariate framework for approximation by max-min neural network operators. Building on the recent advances in approximation theory by neural network operators, particularly, the univariate max-min operators, we propose and analyze new multivariate operators activated by sigmoidal  functions. We establish pointwise and uniform convergence theorems and derive quantitative estimates for the order of approximation via modulus of continuity and multivariate generalized absolute moment. Our results demonstrate that multivariate max-min structure of operators, besides their algebraic elegance, provide efficient and stable approximation tools in both theoretical and applied settings.
\end{abstract} 
\begin{center}
\textbf{1. Introduction and Preliminary}
\end{center}
A neural network consists of simple and highly interconnected processors called neurons. Each neuron stores a  weight $\bar{a}\in \mathbb{R}^r$ and a threshold $b\in \mathbb{R}$ in the memory and evaluates certain activation function $\mu(\bar{a}\cdot \bar{y} + b)$ on receiving an input $\bar{y} \in \mathbb{R}^r$ where $\bar{a} \cdot \bar{y}$ denotes the inner product of $\bar{a}$ and $\bar{y}$ in $\mathbb{R}^r$. These neurons are arranged in layers. A feed forward neural network with multi input, single output and a single hidden layer, whose general scheme is reported in  Figure 1, is mathematically represented by
\[
F_n(\bar{y})=\sum_{i=0}^{n} c_i\,\mu(\bar{a}_i\cdot \bar{y} + b_i),
\]
where $n \in \mathbb{N}$  denotes the number of neurons in the hidden layer 
and $c_i$ are constants to be adjusted in the learning stage, and $\mu$ is a non-linear function called the activation function. Mostly activation functions are sigmoidal functions.\\
\noindent\textbf{Sigmoidal function:} A function $\mu:\mathbb{R}\to\mathbb{R}$ is called sigmoidal iff
\[
\mu(y)=
\begin{cases}
1, & y\to +\infty,\\[0.3em]
0, & y\to -\infty.
\end{cases}
\]
\begin{figure}[htbp]
\centering
\includegraphics[width=0.8\textwidth]{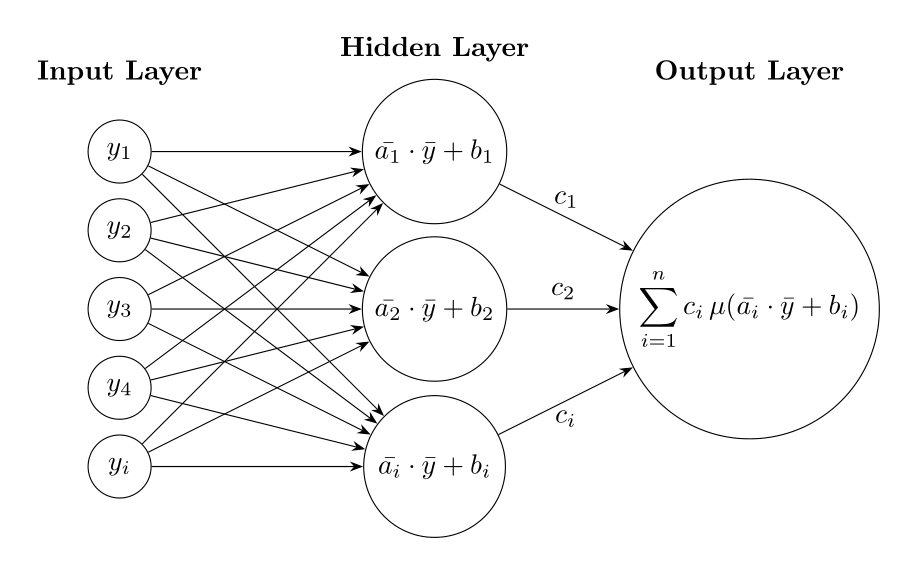}
\caption{Architecture of a single hidden layer feed forward neural network with $n$ hidden neurons.}
\end{figure}
In the literature, various other types of activation functions exist, including ramp functions, binary step functions, and the Rectified Linear Unit (ReLU) function etc. For more details one can  see \cite{ref2}.
    It should be noted that sigmoidal functions are widely studied in neural network and approximation theory (see \cite{ref7,ref8,ref9,ref10,ref11,ref12,ref13,ref14,ref15,ref16,ref17,ref18,ref21}). Anastassiou \cite{ref19} conducted a study on neural network operators defined by Cardaliaguet-Euvrard type \cite{ref20} and derived a rate for the approximation of the NNOs for continuous functions. Subsequently, he investigated neural network operators activated by the logistic function and hyperbolic tangent function in one-dimensional and multi-dimensional cases \cite{ref7,ref8,ref21,ref22,ref23} . 
After that, Costarelli and Spigler \cite{ref10,ref11} examined a unified approach for a general class of sigmoidal functions. Later, Costarelli and Vinti \cite{ref13,ref24} studied the max-product versions of these operators and improved the rate of approximation. Since max-product operators accelerate the rate of approximation, these are extensively used in approximation theory \cite{ref25,ref26,ref27,ref28,ref29,ref30,ref31,ref32,ref33,ref34,ref35,ref36,ref37}.
It was Bede and his colleagues who first obtained a pseudo-linear approach in approximation theory \cite{ref30}. Here, changing the algebraic structure of fields of real numbers with max-product semiring, max-min semiring, and semirings with generated pseudo-operations, the authors obtained pseudo-linear versions of the Shepard operator \cite{ref31}. Moreover, the authors achieved a higher level of approximation accuracy in the max-product case along with reduced computational time complexity in both the max-product and max-min cases \cite{ref50}. We note that although max-min operators are primarily associated with fuzzy theory \cite{ref38,ref39}, because of their approximation capabilities and lower computational complexity, max-min operators have recently begun to be studied in approximation theory \cite{ref40,ref41,ref42,ref43}.\\
In this paper, we use the following notations: \[
\mathcal{R} := [a_1,b_1] \times \cdots \times [a_r, b_r] \subset \mathbb{R}^r, \text{a multivariate box-domain in $\mathbb{R}^r$,}
\]
 
 \[
\bigvee_{i \in J} a_i := \sup\{a_i \in \mathbb{R} : i \in J\},
\]and \[
\bigwedge_{i \in J} b_i := \inf\{b_i\in \mathbb{R} : i \in J\},
\]
where $J$ is any set of indices.  The $r$ dimensional maximum (supremum) operation 
$\;\vee \cdots \vee_{(k_{1},\ldots,k_{r}) \in \mathbb{Z}^{r}}\;$ 
is indicated by ``$\;\vee_{\bar{k} \in \mathbb{Z}^{r}}\;$''.
Clearly, if the cardinality of $J$ is finite, then supremum and infimum in the above definitions reduce 
to a maximum and a minimum, respectively.
\[
\rho_{\mu}(\bar{y}) := \varphi_{\mu}(y_1)\,\varphi_{\mu}(y_2)\cdots 
\varphi_{\mu}(y_r), 
\qquad \bar{y} = (y_1,\ldots,y_r) \in \mathbb{R}^{r}. \tag{1.1}
\] is the multivariate density function, where \[
\varphi_{\mu}(y) := \frac{\mu(y+1) - \mu(y-1)}{2}, 
\qquad y \in \mathbb{R}.\tag{1.2}
\] 
For a given $\bar{y}\in\mathcal{R}$, $n\in\mathbb{N}$ and $\delta>0$, we define 
\[
B_{\delta,n}(\bar{y})
:=
\left\{
\bar{k} \in J_n  :
\left|\left| \frac{\bar{k}}{n} - \bar{y} \right|\right|_2 > \delta
\right\}.
\]
\textbf{Lemma 1.1.}\cite{ref30} The maximum and minimum operations follow the following properties.
\begin{itemize}
\item[($C_1$)] 
If for any $a_k, b_k \in \mathbb{R}$ $(k \in \mathbb{N})$ satisfying
$\displaystyle \bigvee_{k\in\mathbb{N}} a_k < \infty$ or 
$\displaystyle \bigvee_{k\in\mathbb{N}} b_k < \infty$, then we get
\begin{equation}\label{eq2.3}
\left|\; \bigvee_{k\in\mathbb{N}} a_k \;-\; 
\bigvee_{k\in\mathbb{N}} b_k \;\right|
\;\le\;
\bigvee_{k\in\mathbb{N}} |a_k - b_k|.\notag
\end{equation}
Here, we should state that the above lemma is also valid for all $k\in\mathbb{Z}$.
\item[($C_2$)] 
If  $x, y, z \in [0,1]$, then
$|\, x \wedge y - x \wedge z \,|
\;\le\;
x \wedge |\, y - z \,|.$\notag

\end{itemize}
 The neural network operator $F_n$  defined as
\[
F_n(h,\bar{y}):=
\frac{
\displaystyle\sum_{k_1=\lceil na_1\rceil}^{\lfloor nb_1\rfloor}
\cdots
\sum_{k_r=\lceil na_r\rceil}^{\lfloor nb_r\rfloor}
h\left(\frac{\bar{k}}{n}\right)\rho_\mu(n\bar{y}-\bar{k})
}{
\displaystyle\sum_{k_1=\lceil na_1\rceil}^{\lfloor nb_1\rfloor}
\cdots
\sum_{k_r=\lceil na_r\rceil}^{\lfloor nb_r\rfloor}
\rho_\mu(n\bar{y}-\bar{k})
},
\qquad \bar{y}\in \mathcal{R}, \tag{1.3}
\] 
was introduced by Anastassiou \cite{ref22}, where $\lceil \cdot \rceil$ and $\lfloor \cdot \rfloor$ are used to denote the ceiling and integer part, respectively. 
Later in \cite{ref24}, the summation in (1.3) was replace with maximum operation.
In the maximum operation setting, Costarelli and Vinti \cite{ref24} constructed the multivariate max-product 
neural network operator given by
\begin{equation}
F^{(M)}_n(h;\bar{y}):=
\frac{
\displaystyle \bigvee_{\bar{k} \in J^r_n}
\left( h\!\left(\frac{\bar{k}}{n}\right)\rho_\mu(n\bar{y}-\bar{k}) \right)
}{
\displaystyle \bigvee_{\bar{k}\in J^r_n}
\rho_\mu(n\bar{y}-\bar{k})
},
\qquad \bar{y}\in \mathcal{R},\tag{1.4}
\label{eq:maxminop}
\end{equation}
 here $J^r_n=\{\lceil na_1\rceil,\lfloor nb_1\rfloor\} \times \cdots \times \{\lceil na_r\rceil,\lfloor nb_r\rfloor\} \subset \mathbb{R}^r,\bar{k} = 
 (k_1, \cdots ,k_r)$ and the superscript $(M)$ in the 
operator $F^{(M)}_n$ denote the max-product operator. Besides the pointwise and 
uniform convergence theorems of the above operators, the authors in \cite{ref13,ref24} 
also investigated the quasi-interpolation (QI) operators aiming at approximating 
functions on $\mathbb{R}^r$. For this purpose, the indices of the summation and maximum 
operations were replaced with $\mathbb{Z}^r$ to $J^r_n$ in the definitions of the operators 
\((1.3)\) and \((1.4)\).\\
Recently, Aslan  \cite{ref50} has introduced max-min neural network operators defined as: 
\begin{equation}
F^{(m)}_n h(y)
:=
\displaystyle
\bigvee_{k = \lceil{n a}\rceil}^{\lfloor{n b}\rfloor}
\left[
h\!\left(\frac{k}{n}\right)
\wedge\left(
\frac{\varphi_\mu(n y - k)
}{
\displaystyle
\bigvee_{d = \lceil{n a}\rceil}^{\lfloor{n b}\rfloor}
\varphi_\mu(n y - d)
}\right)\right],
\qquad
y \in [a,b],
\tag{1.5}
\end{equation}
where $n \in \mathbb{N}$ is so large that $\lceil n a\rceil \le \lfloor n b\rfloor$.
Here, the superscript $(m)$ of $F^{(m)}_n$ denotes the max-min operator.
This operator  performs better than the classical
neural network operator. However, it is slightly outperformed by the
max-product neural network operator. The computational time and
complexity are almost the same in all cases \cite{ref50}.\\It is intriguing to achieve a multivariate extension of the results demonstrated by Aaslan  \cite{ref50} because neurocomputing processes typically include high dimensional data.  For this reason, in this paper, we construct  neural network operators in the max-min setting for the multivariate type function. 
Furthermore, we demonstrate that our well-define multivariate  max-min neural network operator  is pseudo-linear. We establish both pointwise and uniform convergence 
properties for the operator. Then, using the modulus of continuity and multivariate generalized absolute moment,  we derive a rate of approximation. Also similar results are  obtained for multivariate max-min type 
QI operators. At the end, we present some numerical examples and provide some graphical representation.
\\ \noindent\textbf{Definition 1.2.} Let 
$h:\mathcal{R}\to[0,1]$ be given. Then the multivariate extension of the  operation (1.5), denoted by  $\mathcal{L}_n$, is defined as:
\begin{equation}
\mathcal{L}_n(h;\bar{y}):=\bigvee_{\bar{k} \in J^r_n}
\left[
h\!\left(\frac{\bar{k}}{n}\right)
\wedge\left(
\frac{\rho_\mu(n \bar{y} - \bar{k})
}{
\displaystyle
\bigvee_{\bar{d}\in J^r_n}
\rho_\mu(n\bar{y}-\bar{d})
}\right)\right],
\qquad \bar{y}\in \mathcal{R},\tag{1.6}
\end{equation}
where $n\in\mathbb N$ is so large such that $\lceil na_i\rceil\le \lceil nb_i\rceil$, for all $i\in\{1\dotsc r\}$. \\
Contrary to max-product and classical neural network operators, 
the max-min neural network operator is not homogeneous, that is, for every 
$\alpha\in(0,1)$,
\[
\mathcal{L}_n(\alpha h) \ne \alpha \mathcal{L}_n(h).
\]
In this paper, we consider non-decreasing sigmoidal functions $\mu$ such that $\mu(2)>\mu(1)$ and  
\begin{enumerate}
    \item[(a)] $\mu(y) - \tfrac{1}{2}$ is an odd function;
    \item[(b)] $\mu\in C^{2}(\mathbb{R})$ is concave for $y \ge 0$, where $C^{2}(\mathbb{R})$ denotes
2-times continuously differentiable over $\mathbb{R}$;
    \item[(c)] $\mu(y) = O(|y|^{-\alpha})$ as $y \to -\infty$, for some $\alpha > 0$, that is,  there exist constants $C,L>0$ such that
               $
               \mu(y) \le C\,|y|^{-\alpha},
               \qquad y <- L.$
\end{enumerate}
\noindent\textbf{Definition 1.3.}  
We define  “multivariate generalized absolute 
moment for $\rho_\mu$ of order $\beta>0$’’ given by
\[
    m_\beta(\rho_\mu)
    := \sup_{\bar{y}\in\mathbb{R}^{r}}
        \left[
        \bigvee_{\bar{k}\in\mathbb{Z}^{r}}
            \rho_\mu(\bar{y}-\bar{k})\,\|\bar{y}-\bar{k}\|_{2}^{\beta}
        \right].
\]
The above definition extends the univariate case first introduced in \cite{ref13} to multivariate case. \\
In \cite{ref24}, some important properties of the function $\rho_\mu(\bar{y})$
were proved. We recall a few of these properties in the following lemma.\\
\noindent\textbf{Lemma 1.4.} The following properties holds for $\rho_\mu$ defined in (1.1) for any $\mu$ satisfying all  the assumptions.\begin{itemize}
\item[($A_1$)]$\rho_\mu(\bar{y}) = O(||\bar{y}||_2^{-\alpha})$ as $||y||_2 \to +\infty$, where 
               $\alpha$ is the positive constant of condition (c), i.e., 
               there exist constants $C,L>0$ such that
               \[
               \rho_\mu(\bar{y}) \le C\,||\bar{y}||_2^{-\alpha},
               \qquad ||\bar{y}||_2 > L.
               \]
\item[($A_2$)] For any fixed $\bar{y} \in \mathbb{R}^{r}$, there holds
    \[
        \bigvee_{\bar{k} \in J^r_{n}} \rho_\mu(n\bar{y}-\bar{k}) \ge [\varphi_\mu(1)]^{r} > 0.
    \]
    \item[($A_3$)] For every fixed $\bar{y} \in \mathbb{R}^{r}$, there holds
    \[
        \bigvee_{\bar{k} \in \mathbb{Z}^{r}} \rho_\mu(n\bar{y}-\bar{k}) \ge [\varphi_\mu(1)]^{r} > 0,
        \quad  n \in \mathbb{N}.
    \]
\item[($A_4$)]
For every $\gamma > 0$, we have
\[
    \bigvee_{\substack{\bar{k} \in \mathbb{Z}^{r} \\ \|\bar{y}-\bar{k}\|_{2} > {n\gamma}}}
    \rho_\mu(\bar{y}-\bar{k}) = O(n^{-\alpha}), \qquad \text{as } n \to +\infty,
\]
uniformly with respect to $\bar{y} \in \mathbb{R}^{r}.$
\item[($A_5$)]
For every $0 \le \beta \le \alpha$, there holds $m_{\beta}(\rho_\mu) < +\infty$,
where $\alpha$ is the constant of condition (c).
In particular, it turns out that
\[
    m_{0}(\rho_\mu) \le [\varphi_{\mu}(0)]^{r} \le 2^{-r}.
\]
\end{itemize}
Since $h$ is bounded, by Lemma 1.4 ($A_2$), we can see that 
$\mathcal{L}_n$ is well-defined.\\
\begin{center}
\large{\textbf{2. Convergence of multivariate max-min neural network operators}}\\
\end{center}
In this section, we prove the following approximation theorem by means of the operator $\mathcal{L}_n$ defined in (1.6).\\
\textbf{Theorem 2.1.}
Let $h:\mathcal{R}\to[0,1]$ be continuous at some $\bar{y}_0\in\mathcal{R}$. Then,
\[
\lim_{n\to\infty} \mathcal{L}_n(h;\bar{y}_0) = h(\bar{y}_0).
\] Moreover, if $h\in C(\mathcal{R},[0,1])$, 
then the convergence is uniform, that is,
\[
\lim_{n\to\infty} \| \mathcal{L}_n(h) - h \| = 0,
\]
where $\|\cdot\|$ denotes the supremum norm. We need the following lemma for proving our theorem. 
\\
\textbf{Lemma 2.2.}
Let $h,g : \mathcal{R} \to [0,1]$ be two functions. Then,
\begin{itemize}
\item[($B_1$)] $\mathcal{L}_n(h)$ is continuous on $\mathcal{R}$ for all  $\mu$ continuous on $\mathbb{R}$.
\item[($B_2$)]
If $h(\bar{y}) \le g(\bar{y})$ for every $\bar{y}\in 
\mathcal{R}$, then 
$\mathcal{L}_n(h;\bar{y}) \le \mathcal{L}_n(g;\bar{y})$ for all $\bar{y}\in \mathcal{R}$.
\item[($B_3$)]
$\mathcal{L}_n$ is pseudo-linear in the max-min sense, that is,
\[
\mathcal{L}_n\!\left( (\alpha\wedge h)\, \vee\, (\beta\wedge g) ; \bar{y} \right)
  = \alpha \wedge \mathcal{L}_n(h;\bar{y})\ \vee\ 
    \beta \wedge \mathcal{L}_n(g;\bar{y}).
\]
for all $h,g : Y\subset \mathbb{R}^r \to [0,1]$ and for every $\alpha,\beta\in [0,1]$.
\item[($B_4$)]
$\mathcal{L}_n$ is subadditive, that is,
\[
\mathcal{L}_n(h+g;\bar{y})
\le
\mathcal{L}_n(h;\bar{y}) + \mathcal{L}_n(g;\bar{y}),
\qquad\text{for all } \bar{y}\in\mathcal{R}.
\]
\item[($B_5$)]
$\bigl|\mathcal{L}_n(h;\bar{y}) -\mathcal{L}_n(g;\bar{y}) \bigr|
\le 
\mathcal{L}_n(|h-g|;\bar{y})$ on $\mathcal{R}$,
for sufficiently large $n\in\mathbb{N}$.
\end{itemize}
\textbf{Proof.}
For ($B_1$),
given that  $\mu$ is continuous on $\mathbb{R}$, so the function
\[
\varphi_\mu(y)=\frac{\mu(y+1)-\mu(y-1)}{2}
\]
is also continuous on $\mathbb{R}$ and so $\rho_\mu$. Hence,
\[\frac{\rho_\mu(n \bar{y} - \bar{k})
}{
\displaystyle
\bigvee_{\bar{d}\in J^r_n}
\rho_\mu(n\bar{y}-\bar{d})
} \text{ is continuous. So }
 h\!\left(\tfrac{\bar{k}}{n}\right)\wedge \left(
\frac{\rho_\mu(n \bar{y} - \bar{k})
}{
\displaystyle
\bigvee_{\bar{d}\in J^r_n}
\rho_\mu(n\bar{y}-\bar{d})
}\right)
\]
is continuous. Since 
maximum of finitely many continuous functions is continuous, the
operator $\mathcal{L}_n(h)$ is continuous on $\mathcal{R}$.
\\For ($B_2$), let $\bar{y}\in\mathcal{R}$ be fixed and given that $h(\bar{y})\le g(\bar{y})$ for all
$\bar{y}\in\mathcal{R}$. So, for every $\bar{k}$,
\[
h\!\left(\tfrac{\bar{k}}{n}\right)\wedge \left(
\frac{\rho_\mu(n \bar{y} - \bar{k})
}{
\displaystyle
\bigvee_{\bar{d}\in J^r_n}
\rho_\mu(n\bar{y}-\bar{d})
}\right)
\le
g\!\left(\tfrac{\bar{k}}{n}\right)\wedge \left(
\frac{\rho_\mu(n \bar{y} - \bar{k})
}{
\displaystyle
\bigvee_{\bar{d}\in J^r_n}
\rho_\mu(n\bar{y}-\bar{d})
}\right),
\]
since $\frac{\rho_\mu(n \bar{y} - \bar{k})
}{
\displaystyle
\bigvee_{\bar{d}\in J^r_n}
\rho_\mu(n\bar{y}-\bar{d})
}>0$  and the minimum operation is monotone. Taking the maximum over
$\bar{k}\in J^r_n$, we obtain
$\mathcal{L}_n(h;\bar{y})\le \mathcal{L}_n(g;\bar{y}),
\quad \forall  \bar{y}\in\mathcal{R}.
$\\
For ($B_3$), let $\bar{y}\in \mathcal{R}$ be fixed and $\alpha,\beta\in[0,1]$.
Then since $(\cdot,\wedge,\vee)$ is an ordered semiring, there holds\[
\begin{aligned}
\mathcal{L}_n\!\left( (\alpha\wedge h)\vee(\beta\wedge g); \bar{y} \right)
&=
\bigvee_{\bar{k}\in J^r_n}
\left[
(\alpha\wedge h)\!\left(\frac{\bar{k}}{n}\right)
\ \vee\
(\beta\wedge g)\!\left(\frac{\bar{k}}{n}\right)
\right]
\wedge
\left(
\frac{\rho_\mu(n \bar{y} - \bar{k})
}{
\displaystyle
\bigvee_{\bar{d}\in J^r_n}
\rho_\mu(n\bar{y}-\bar{d})
}\right)
\\
&=
\bigvee_{\bar{k}\in J^r_n}
\left[
  \left(
  \alpha\wedge h\!\left(\frac{\bar{k}}{n}\right)
  \right)
  \wedge
  \left(
\frac{\rho_\mu(n \bar{y} - \bar{k})
}{
\displaystyle
\bigvee_{\bar{d}\in J^r_n}
\rho_\mu(n\bar{y}-\bar{d})
}\right)
\right]
\\
&\qquad\qquad\vee
\bigvee_{\bar{k}\in J^r_n}
\left[
  \left(
  \beta\wedge g\!\left(\frac{\bar{k}}{n}\right)
  \right)
  \wedge
  \left(
\frac{\rho_\mu(n \bar{y} - \bar{k})
}{
\displaystyle
\bigvee_{\bar{d}\in J^r_n}
\rho_\mu(n\bar{y}-\bar{d})
}\right)\right]
\\
&=
\alpha\wedge
\bigvee_{\bar{k}\in J^r_n}
\left[
h\!\left(\frac{\bar{k}}{n}\right)
\wedge
\left(
\frac{\rho_\mu(n \bar{y} - \bar{k})
}{
\displaystyle
\bigvee_{\bar{d}\in J^r_n}
\rho_\mu(n\bar{y}-\bar{d})
}\right)\right]
\\
&\qquad\vee
\beta\wedge
\bigvee_{\bar{k}\in J^r_n}
\left[
g\!\left(\frac{\bar{k}}{n}\right)
\wedge
\left(
\frac{\rho_\mu(n \bar{y} - \bar{k})
}{
\displaystyle
\bigvee_{\bar{d}\in J^r_n}
\rho_\mu(n\bar{y}-\bar{d})
}\right)\right]
\\
&=
\alpha\wedge \mathcal{L}_n(h;\bar{y})
\ \vee\
\beta\wedge \mathcal{L}_n(g;\bar{y}),
\end{aligned}
\]
for sufficiently large $n\in\mathbb{N}$.
The proof of part ($B_4$) can  be derived trivially from the inequality
\[
a\wedge(b+c) \le a\wedge b + a\wedge c, 
\qquad\text{for all } a,b,c\ge 0.
\]
For the proof of ($B_5$), from Lemmas 1.1, we have
\[
\begin{aligned}
|\mathcal{L}_n(h;\bar{y}) - \mathcal{L}_n(g;\bar{y})|
&\le
\bigvee_{\bar{k}\in J^r_n}\left|
\left[
  h\!\left(\frac{\bar{k}}{n}\right)
  \wedge
  \left(
\frac{\rho_\mu(n \bar{y} - \bar{k})
}{
\displaystyle
\bigvee_{\bar{d}\in J^r_n}
\rho_\mu(n\bar{y}-\bar{d})
}\right)\right]-
\left[g\!\left(\frac{\bar{k}}{n}\right)
\wedge
\left(
\frac{\rho_\mu(n \bar{y} - \bar{k})
}{
\displaystyle
\bigvee_{\bar{d}\in J^r_n}
\rho_\mu(n\bar{y}-\bar{d})
}\right)\right]\right|
\\
&\le
\bigvee_{\bar{k}\in J^r_n}
\left[
  \left| h\!\left(\frac{\bar{k}}{n}\right) - g\!\left(\frac{\bar{k}}{n}\right) \right|
  \wedge
  \left(
\frac{\rho_\mu(n \bar{y} - \bar{k})
}{
\displaystyle
\bigvee_{\bar{d}\in J^r_n}
\rho_\mu(n\bar{y}-\bar{d})
}\right)\right]
\\
&=
\mathcal{L}_n(|h-g|;\bar{y}).
\end{aligned}
\]
\textbf{Proof of Theorem 2.1.}
Let $h:\mathcal{R}\to[0,1]$ be   continuous at $\bar{y}_0\in\mathcal{R}$.
Then, by using   triangle inequality, we have\[
\begin{aligned}
\left|\mathcal{L}_n(h;\bar{y}_0) - h(\bar{y}_0)\right|
&=
\left|\bigvee_{\bar{k}\in J^r_n}
\left[
  h\!\left(\frac{\bar{k}}{n}\right)
  \wedge
  \left(
\frac{\rho_\mu(n \bar{y} - \bar{k})
}{
\displaystyle
\bigvee_{\bar{d}\in J^r_n}
\rho_\mu(n\bar{y}-\bar{d})
}\right)\right]
- h(\bar{y}_0)
\right|
\\
&\le
\left|
\bigvee_{\bar{k}\in J^r_n}
\left[
  h\!\left(\frac{\bar{k}}{n}\right)
  \wedge
  \left(
\frac{\rho_\mu(n \bar{y} - \bar{k})
}{
\displaystyle
\bigvee_{\bar{d}\in J^r_n}
\rho_\mu(n\bar{y}-\bar{d})
}\right)\right]
\right.
\\
&\qquad\qquad
\left.
-
\bigvee_{\bar{k}\in J^r_n}
\left[
  h(\bar{y}_0)
  \wedge
  \left(
\frac{\rho_\mu(n \bar{y} - \bar{k})
}{
\displaystyle
\bigvee_{\bar{d}\in J^r_n}
\rho_\mu(n\bar{y}-\bar{d})
}\right)\right]
\right|
\\
&\qquad
+
\left|
\bigvee_{\bar{k}\in J^r_n}
\left[
  h(\bar{y}_0)
  \wedge
  \left(
\frac{\rho_\mu(n \bar{y} - \bar{k})
}{
\displaystyle
\bigvee_{\bar{d}\in J^r_n}
\rho_\mu(n\bar{y}-\bar{d})
}\right)\right]
- h(\bar{y}_0)
\right|.
\end{aligned}
\]
This holds for sufficiently large $n\in\mathbb{N}$.  
In the second expression, one can easily sees, by the pseudo-linearity of $\mathcal{L}_n$, that
\[
\begin{aligned}
\left|
\bigvee_{\bar{k}\in J^r_n}
\left[
h(\bar{y}_0)\wedge 
\left(
\frac{\rho_\mu(n \bar{y} - \bar{k})
}{
\displaystyle
\bigvee_{\bar{d}\in J^r_n}
\rho_\mu(n\bar{y}-\bar{d})
}\right)\right]
- h(\bar{y}_0)
\right|
&=
\left|
h(\bar{y}_0)\wedge
\left(
\frac{\bigvee_{\bar{k}\in J^r_n}\rho_\mu(n \bar{y} - \bar{k})
}{
\displaystyle
\bigvee_{\bar{d}\in J^r_n}
\rho_\mu(n\bar{y}-\bar{d})
}\right)
\right|
\\
&=
\left|
h(\bar{y}_0)\wedge 1 - h(\bar{y}_0)
\right|
=0.
\end{aligned}
\]
Then, from Lemmas 1.1, we have
\begin{align}
\left|\mathcal{L}_n(h;\bar{y}_0) - h(\bar{y}_0)\right| \notag
&\le
\bigvee_{\bar{k} \in J^r_n}
\left|
  h\!\left(\frac{\bar{k}}{n}\right)\wedge 
  \left(
\frac{\rho_\mu(n \bar{y} - \bar{k})
}{
\displaystyle
\bigvee_{\bar{d}\in J^r_n}
\rho_\mu(n\bar{y}-\bar{d})
}\right)
  \ -\ 
  h(y_0)\wedge 
  \left(
\frac{\rho_\mu(n \bar{y} - \bar{k})
}{
\displaystyle
\bigvee_{\bar{d}\in J^r_n}
\rho_\mu(n\bar{y}-\bar{d})
}\right)
\right|
\\ \notag
&\le
\bigvee_{\bar{k}\in J^r_n}
\left|
  h\!\left(\frac{\bar{k}}{n}\right) - h(\bar{y}_0)
\right|
\wedge
\left(
\frac{\rho_\mu(n \bar{y} - \bar{k})
}{
\displaystyle
\bigvee_{\bar{d}\in J^r_n}
\rho_\mu(n\bar{y}-\bar{d})
}\right).\tag{1.7}
\end{align}
This holds for sufficiently large $n\in\mathbb{N}$.  
Since $h$ is continuous at $\bar{y}_0$, then for any $\varepsilon>0$ there exists  
a $\delta=\delta(\varepsilon,\bar{y}_0)$ such that $||\bar{x}-\bar{y}_0||_2 <\delta$ for all  $\bar{x}\in\mathcal{R}$, 
\[
\left| h(\bar{y}_0)-h(\bar{x})\right|<\varepsilon. \tag{1.8}
\]
Splitting the max operation $\bigvee_{\bar{k}\in J^r_n}$ into $\bigvee_{\bar{k}\in B_{\delta,n}(\bar{y}_0)} \vee \bigvee_{\bar{k}\notin B_{\delta,n}(\bar{y}_0)}$, we have
\[
\begin{aligned}
\left|\mathcal{L}_n(h;\bar{y}_0)-h(\bar{y}_0)\right|
&\le
\bigvee_{\bar{k}\notin B_{\delta,n}(\bar{y}_0)}
\left|
  h\!\left(\frac{\bar{k}}{n}\right)-h(\bar{y}_0)
\right|
\wedge
\left(
\frac{\rho_\mu(n \bar{y} - \bar{k})
}{
\displaystyle
\bigvee_{\bar{d}\in J^r_n}
\rho_\mu(n\bar{y}-\bar{d})
}\right)
\\
&\qquad\vee
\bigvee_{\bar{k}\in B_{\delta,n}(\bar{y}_0)}
\left|
  h\!\left(\frac{\bar{k}}{n}\right)-h(\bar{y}_0)
\right|
\wedge
\left(
\frac{\rho_\mu(n \bar{y} - \bar{k})
}{
\displaystyle
\bigvee_{\bar{d}\in J^r_n}
\rho_\mu(n\bar{y}-\bar{d})
}\right)
\\
&= : I_{B^c_{\delta,n}} \vee I_{B_{\delta,n}}.
\end{aligned}
\]
From (1.8), we obtain
\[
I_{B^c_{\delta,n}}
<
\bigvee_{\bar{k}\notin B_{\delta,n}(\bar{y}_0)}
\varepsilon\wedge
\left(
\frac{\rho_\mu(n \bar{y} - \bar{k})
}{
\displaystyle
\bigvee_{\bar{d}\in J^r_n}
\rho_\mu(n\bar{y}-\bar{d})
}\right)
\le \varepsilon.
\]
On the other hand, since $a\wedge b\le b$ for all $a,b\in\mathbb{R}$, it follows from Lemmas 1.4($A_2$) and 1.4($A_4$), that
\[
I_{B_{\delta,n}}\le\bigvee_{\bar{k}\in B_{\delta,n}(\bar{y}_0)}\left|h\!\left(\frac{\bar{k}}{n}\right)-h(\bar{y}_0)\right|\wedge 
\frac{\rho_\mu(n\bar{y}_0-\bar{k})}{(\varphi_\mu(1))^r}
\]
\[
\le \frac{1}{(\varphi_\mu(1))^r}\bigvee_{||n\bar{y}_0-\bar{k}||_2> n\delta}\ \rho_\mu(n\bar{y}_0-\bar{k})
\]
\[
\le \frac{K}{(\varphi_\mu(1))^r}\,n^{-\alpha}<\varepsilon,
\]
for sufficiently large $n\in\mathbb{N}$, where $\alpha$ corresponds to the condition $(c)$.\\
Given that the estimate in Lemma 1.4($A_4$) is uniform in $\bar{y}$, we can apply it to $n\bar{y}_0$ instead of $\bar{y}_0$, so that
\[
\big|\mathcal{L}_n(h;\bar{y}_0)-h(\bar{y}_0)\big| < \varepsilon \ \vee\ \varepsilon=\varepsilon
\]
for sufficiently large $n\in\mathbb{N}$.  
Finally, if $h\in C(\mathcal{R},[0,1])$, then replacing $\delta=\delta(\varepsilon, \bar{y})$ with $\delta=\delta(\varepsilon)$ in the proof, we obtain the uniform approximation. 
\\ \noindent\textbf{Remark 2.3.}\cite{ref50} 
Although it is true for multivariate max-min neural network operator using functions whose ranges are in $[0,1]$, it is also possible to generalize the approximation for all bounded functions whose ranges are greater than 1, or negative numbers with the following operators.\\  
Let $h:\mathcal{R}\to\mathbb{R}$ be a bounded function. Then, defining the $\tilde{\mathcal{L}_n}$ as given below,
\[
\tilde{\mathcal{L}_n}(h;\bar{y}):=
\begin{cases}
\mathcal{L}_n(h;\bar{y}), & \text{if } \bar{y}\in h^{-1}[0,1] \\
\big(\mathcal{L}_n(h^{-1};\bar{y})\big)^{-1}, & \text{if }\bar{y}\in h^{-1}(1,\infty)\\
-\,\mathcal{L}_n(-h;\bar{y}), & \text{if } \bar{y}\in h^{-1}[-1,0)\\
-\big(\mathcal{L}_n((-h)^{-1};\bar{y})\big)^{-1}, & \text{if } \bar{y}\in h^{-1}(-\infty,1)
\end{cases}
\]
we can obtain a general approximation result for all bounded functions defined on $\mathcal{R}$.
\begin{center}
\large{\textbf{{3. Rate of approximation for multivariate  max-min neural network operators}}}
\end{center}
In this section, we will investigate the rate of approximation for $h\in C[\mathcal{R},[0,1])$ by using the modulus of continuity of $h$.
For $\delta>0$, the function
\[
\omega_{\mathcal{R}}(h,\delta):=\sup_{||\bar{x}-\bar{y}||_2\le \delta}\{|h(\bar{x})-h(\bar{y})|:\bar{x},\bar{y}\in\mathcal{R}\}
\]
is called the modulus of continuity for  $h\in C[\mathcal{R},[0,1])$ .
If $h$ is defined on $\mathbb{R}^r$, then the modulus of continuity of $h$ is given by
\[
\omega(h,\delta):=\sup_{||\bar{x}-\bar{y}||_2\le\delta}\{|h(\bar{x})-h(\bar{y})|:\bar{x},\bar{y}\in\mathbb{R}^r\}.
\]

Now, we are able to compute the rate of approximation as follows:
\\ \textbf{Theorem 3.1.}  
Let  $\delta_n$ be a null sequence of positive real numbers such that $(n\delta_n)^{-1}$ is also a null sequence. Then, for any $h\in C(\mathcal{R},[0,1])$, we have
\[
\left|\mathcal{L}_n(h)-h\right|\le \omega_{\mathcal{R}}(h,\delta_n)\bigvee{m_\alpha(\rho_\mu)}\,
\frac{1}{(\varphi_\mu(1))^r}\,\frac{1}{n^{\alpha}(\delta_n)^{\alpha}},
\] where $\alpha>0$ correspond to the condition (c).\\
\textbf{Proof.}  
From (1.7), we can write that
\[
\big|\mathcal{L}_n(h;\bar{y})-h(\bar{y})\big|
\le \bigvee_{\bar{k}\in J^r_n }\left|h\!\left(\frac{\bar{k}}{n}\right)-h(\bar{y})\right|\wedge
\left(
\frac{\rho_\mu(n \bar{y} - \bar{k})
}{
\displaystyle
\bigvee_{\bar{d}\in J^r_n}
\rho_\mu(n\bar{y}-\bar{d})
}\right).
\]
\ \ \ \ \
\[
= \bigvee_{\bar{k}\notin B_{\delta_{n,n}}(\bar{y})}\left|h\!\left(\frac{\bar{k}}{n}\right)-h(\bar{y})\right|\wedge
\left(
\frac{\rho_\mu(n \bar{y} - \bar{k})
}{
\displaystyle
\bigvee_{\bar{d}\in J^r_n}
\rho_\mu(n\bar{y}-\bar{d})
}\right)
\]
\[
\ \ \ \ \ \bigvee\ 
\bigvee_{\bar{k}\in B_{\delta_{n,n}}(\bar{y})}\left|h\!\left(\frac{\bar{k}}{n}\right)-h(\bar{y})\right|\wedge
\left(
\frac{\rho_\mu(n \bar{y} - \bar{k})
}{
\displaystyle
\bigvee_{\bar{d}\in J^r_n}
\rho_\mu(n\bar{y}-\bar{d})
}\right)
=:D_{B^c_{\delta_{n,n}}}\vee D_{B_{\delta_{n,n}}}.
\]
Using the modulus of continuity, we have\[D_{B^c_{\delta_{n,n}}}\le \bigvee_{\bar{k}\notin B_{\delta_{n,n}}(\bar{y})}\omega_{\mathcal{R}}\!\left(h,{\bigg|\bigg|\frac{\bar{k}}{n}-\bar{y}\bigg|\bigg|_2}\right)\wedge
\left(
\frac{\rho_\mu(n \bar{y} - \bar{k})
}{
\displaystyle
\bigvee_{\bar{d}\in J^r_n}
\rho_\mu(n\bar{y}-\bar{d})
}\right)
\]
\[ \le \bigvee_{\bar{k}\notin B_{\delta_{n,n}}(\bar{y})} \omega_{\mathcal{R}}(h,\delta_n)\wedge 
\left(
\frac{\rho_\mu(n \bar{y} - \bar{k})
}{
\displaystyle
\bigvee_{\bar{d}\in J^r_n}
\rho_\mu(n\bar{y}-\bar{d})
}\right).
\]
On the other hand, since $h\in[0,1]$, we have 
\[
D_{B_{\delta_{n,n}}} \le \bigvee_{\bar{k}\in B_{\delta_{n,n}}(\bar{y})} 
\frac{\rho_\mu(n\bar{y}-\bar{k})}{
\bigvee_{\bar{d}\in J^r_n}\rho_\mu(n\bar{y}-\bar{d})}
\]
\[
\le \frac{1}{(\varphi_\mu(1))^r}\bigvee_{||n\bar{y}-\bar{k}||_2>n\delta_n}\rho_\mu(n\bar{y}-\bar{k}).
\]
Since $\bar{k}\in B_{\delta_{n,n}}(\bar{y})$, we get
\[
\frac{\big|\big| \frac{\bar{k}}{n}-\bar{y} \big|\big|_2^\alpha}{\delta_n^\alpha}
= \frac{||n\bar{y}-\bar{k}||_2^\alpha}{n^\alpha\delta_n^\alpha}>1,
\quad\text{where, $\alpha$ corresponds to condition (c).}
\]
Therefore, we have
\[
D_{B_{\delta_{n,n}}} <\frac{1}{(\varphi_\mu(1))^r\, n^\alpha(\delta_n)^\alpha}
\bigvee_{||n\bar{y}-\bar{k}||_2>n\delta_n}\rho_\mu(n\bar{y}-\bar{k})\,||n\bar{y}-\bar{k}||_2^\alpha
\]
\[
< \frac{1}{(\varphi_\mu(1))^r\, n^\alpha(\delta_n)^\alpha}
\bigvee_{\bar{k}\in\mathbb{Z}^r}\rho_\mu(n\bar{y}-\bar{k})\,||n\bar{y}-\bar{k}||_2^\alpha
\]
\[
= \frac{m_\alpha(\rho_\mu)}{(\varphi_\mu(1))^r\, n^\alpha(\delta_n)^\alpha},
\]
which completes the proof. Here one can easily see that $m_\alpha(\rho_\mu)$ is finite from Lemma 1.4 ($A_5$). 
\\As a special case, taking Hölder continuous functions of order ${\beta}$ into consideration, we also have the following rates of approximation. For a given $\beta\in(0,1]$, the space $Lip_{\mathcal{R}}(\beta)$ is defined by
\[Lip_{\mathcal{R}}(\beta):=\left\{h\in C(\mathcal{R},[0,1]):
|h(\bar{x})-h(\bar{y})|\le M\,||\bar{x}-\bar{y}||_2^\beta\  \text{for all } \bar{x},\bar{y}\in \mathcal{R} \text{ and for some } M>0\right\}\]
\textbf{Corollary 3.2.}  
Let $h\in Lip_\mathcal{R}(\beta)$ and $\alpha>0$ correspond to the condition (c). Then
\[
\left\|\mathcal{L}_n(h)-h\right\| = O\left(n^{-\frac{\alpha\beta}{\alpha+\beta}}\right)
\quad \text{as } n\to\infty.
\]
It can be easily proved  by taking $\delta_n=n^{-\frac{\alpha}{\alpha+\beta}}$ in the above proof. 
\begin{center}
\large{\textbf{4. Multivariate max-min quasi-interpolation neural network operators}}
\end{center}
In this section, we investigate quasi-interpolation neural network operators in the multivariate max-min setting.  
To achieve these approximations on the whole $\mathbb{R}^r$, the authors in \cite{ref10,ref13} take into account the quasi-interpolation neural network operators, wherein  the previously existing results of \cite{ref21} are improved.  
Now, we deal with the max-min concept in the construction of the operator.
Let $h:\mathbb{R}^r\to[0,1]$ be a bounded function. Multivariate max-min quasi-interpolation neural network operators is defined by
\[
\mathcal{P}_n(h;\bar{y}):=\bigvee_{\bar{k} \in \mathbb{Z}^r}
\left[
h\!\left(\frac{\bar{k}}{n}\right)
\wedge\left(
\frac{\rho_\mu(n \bar{y} - \bar{k})
}{
\displaystyle
\bigvee_{\bar{d}\in  \mathbb{Z}^r}
\rho_\mu(n\bar{y}-\bar{d})
}\right)\right],
\qquad \bar{y}\in \mathcal{R}.\tag{1.9}
\]
Then one can easily see from Lemma 1.4 $(A_3)$ that (1.9) is well-defined for all $n\in\mathbb{N}$.  
In addition, operator (1.9) satisfies all the properties of Lemma 2.2 on $\mathbb{R}^r$.
We denote by $UC(\mathbb{R}^r,[0,1])$, the space of all uniformly continuous functions $h:\mathbb{R}^r\to[0,1]$.  
Our approximation theorem is  as follows:\\
\textbf{Theorem 4.1.}  
Let $h:\mathbb{R}^r\to[0,1]$. Then,
\[
\lim_{n\to\infty} \mathcal{P}_n(h;\bar{y}_0)=h(\bar{y}_0)
\]
at any point $\bar{y}_0\in\mathbb{R}^r$ of continuity of $h$. Moreover, if $h\in UC(\mathbb{R}^r,[0,1])$, then the approximation is uniform, that is,
\[
\lim_{n\to\infty}\left\| \mathcal{P}_n(h)-h\right\| = 0.
\]
\textbf{Remark 4.2.}  
By following the procedures outlined in Remark 2.3, multivariate max-min quasi-interpolation neural network operators can also be extended for all bounded functions on $\mathbb{R}^r$.\\
By using the modulus of continuity of $h$ and multivariate generalized absolute moment
for $\rho_\mu$, we get the following rate of approximation:\\
\textbf{Theorem 4.3.}  
Let  $\delta_n$ be a null sequence of positive real numbers such that $\frac{1}{(n\delta_n)}$ is a null sequence. Then, for any $h\in UC(\mathbb{R}^r,[0,1])$, we have also 
\[
\left\| \mathcal{P}_n(h)-h\right\|\le \omega(h,\delta_n)\bigvee{\frac{m_\alpha(\rho_\mu)}{(\varphi_\mu(1))^r}}\,
\frac{1}{n^\alpha(\delta_n)^\alpha},
\] where $\alpha>0$ correspond to the condition $(c)$.\\
\textbf{Corollary 4.4.}  
Let $h \in Lip(\beta)$ and $\alpha>0$ correspond to the condition $(c)$.  
Then
\[
\left\| \mathcal{P}_n(h) - h\right\|
= O\!\left(n^{-\frac{\alpha\beta}{\alpha+\beta}}\right)
\qquad \text{as } n \to \infty.
\]
We remark that proofs of Theorems 4.1, 4.3, and Corollary 4.4 can be obtained  
by following the procedure of proof of Theorems 2.1, 3.1, and Corollary 3.2,  respectively.\\
\textbf{Remark 4.5.}
Note that condition (b) is sufficient only. There are examples of $\mu$ which do not  satisfy 
(b), but satisfy ($A_2$) and ($A_3$).
\begin{center}
\large{\textbf{5. Numerical Examples}}
\end{center}
In this section, we present some examples of sigmoidal functions that satisfy our results. We also discuss certain sigmoidal functions that do not satisfy all of our assumptions but still work for our results.
The logistic function and the hyperbolic tangent function, which satisfy our assumptions, are defined, respectively, as follows:
\[
\mu_{\ell}(y):=\frac{1}{(1+e^{-y})},\qquad
\mu_{h}(y):=\frac{\tanh(y)+1}{2},\qquad y\in\mathbb{R}.
\]
Trivially, $\mu_\ell$ and $\mu_h$ are $C^2(\mathbb{R})$ functions and satisfy
assumptions (a), (b), and (c). In particular, due to their exponential decay
to zero as $y \to -\infty$,  condition (c) is satisfied for every $\alpha > 0$.
\\An example of a continuous sigmoidal function which is  not differentiable on $\mathbb{R}$, in particular, which does   not satisfy
assumption (b), is the
well-known ramp function $\mu_R$, defined as follows:

\[
\mu_R(y):=
\begin{cases}
0, & y < -\frac{1}{2},\\[4pt]
y + \frac{1}{2}, & -\frac{1}{2} \le y \le \frac{1}{2},\\[4pt]
1, & y > \frac{1}{2}.
\end{cases}
\] The corresponding $\varphi_{\mu_R}$ has compact support 
$[-1,1]$, then also $\rho_{\mu_R}$ has compact support $[-1,1] \times
\cdots \times [-1,1]$. \\ An example of a discontinuous (step) sigmoidal function that satisfies
assumptions (a) and (c), but not (b), is the following:
\[
\mu_{three}(y):=
\begin{cases}
0, & y < -1/2,\\[4pt]
1/2, & -1/2 \le y \le 1/2,\\[4pt]
1, & y > 1/2.
\end{cases}
\]
All the above presented examples satisfy condition $(c)$
for $\alpha \ge 1$, then the corresponding multivariate neural network operators  and quasi-interpolation
operators achieve the rate of approximation established in
Theorems 3.1 and 4.3, respectively.\\ Now, we construct
examples of continuous functions but not differentiable sigmoidal functions which satisfy assumption $(c)$ for
$0 < \alpha < 1$, we can define as:
\[
\mu_\gamma(y):=
\begin{cases}
\displaystyle \frac{1}{|y|^\gamma + 2}, & x < -2^{1/\gamma},\\[8pt]
2^{-(1/\gamma)-2} x + (1/2), & -2^{1/\gamma} \le x \le 2^{1/\gamma},\\[8pt]
\displaystyle \frac{x^\gamma + 1}{y^\gamma + 2}, & y > 2^{1/\gamma},
\end{cases}
\]
where $\gamma > 0$. Through simple computations, it can be proved that
$\mu_\gamma$ are continuous sigmoidal functions such that condition $(c)$ is
satisfied for $0 < \alpha \le \gamma$. By using $\mu_\gamma$, we can construct    $\varphi_{\mu_{\gamma}}$ and $\rho_{\mu_{\gamma}}$ by using (1.2) and (1.1), respectively.\\
Now, we demonstrate the approximation ability of the multivariate max-min neural network operator by comparing its approximation errors with multivariate classical and multivariate max-product NN operators activated by $\mu_l$.
Let 
\[
h : [0,1] \times [0,1] \to [0,1]
\]
be defined by
\[
h(y_1, y_2) =  \frac{y_1^2+y_2^2}{2}.
\]
From the data shown in Table 1, it is clear that the multivariate max-min neural network operator performs better than the classical multivariate neural network operators.
\begin{table}[h!]
\centering
\renewcommand{\arraystretch}{1.4} 
\caption{Comparison of approximation errors between multivariate classical, multivariate max-product and multivariate max-min neural network operators.}
\vspace{0.25cm}
\begin{tabular}{c c c c}
\hline
$n$ 
& $\|F_n(f)-f\|$ 
& $\|F_n^{(M)}(h)-h\|$ 
& $\|\mathcal{L}_n(h)-h\|$ \\
\hline
20   & 0.10867 & 0.043789 & 0.096539 \\
55   & 0.041218 & 0.016589 & 0.037195 \\
77   & 0.02964 & 0.011673 & 0.026791 \\
100  & 0.022911 & 0.0066111 & 0.020389 \\
150  & 0.015339 & 0.00010367 & 0.013133 \\
1000 & 0.0023175 & 0.00066211 & 0.0020606 \\
\hline
\end{tabular}
\end{table}
\\As is evident from the data presented in Table 1, the multivariate max-min neural network operator outperforms the classical multivariate neural network operator.
On the other hand, it is slightly outperformed by the multivariate max-product neural network operator.
The computational time and complexity are almost the same in all cases.
\begin{figure}[ht]
\centering
\begin{minipage}{0.48\textwidth}
\centering
\includegraphics[width=\textwidth]{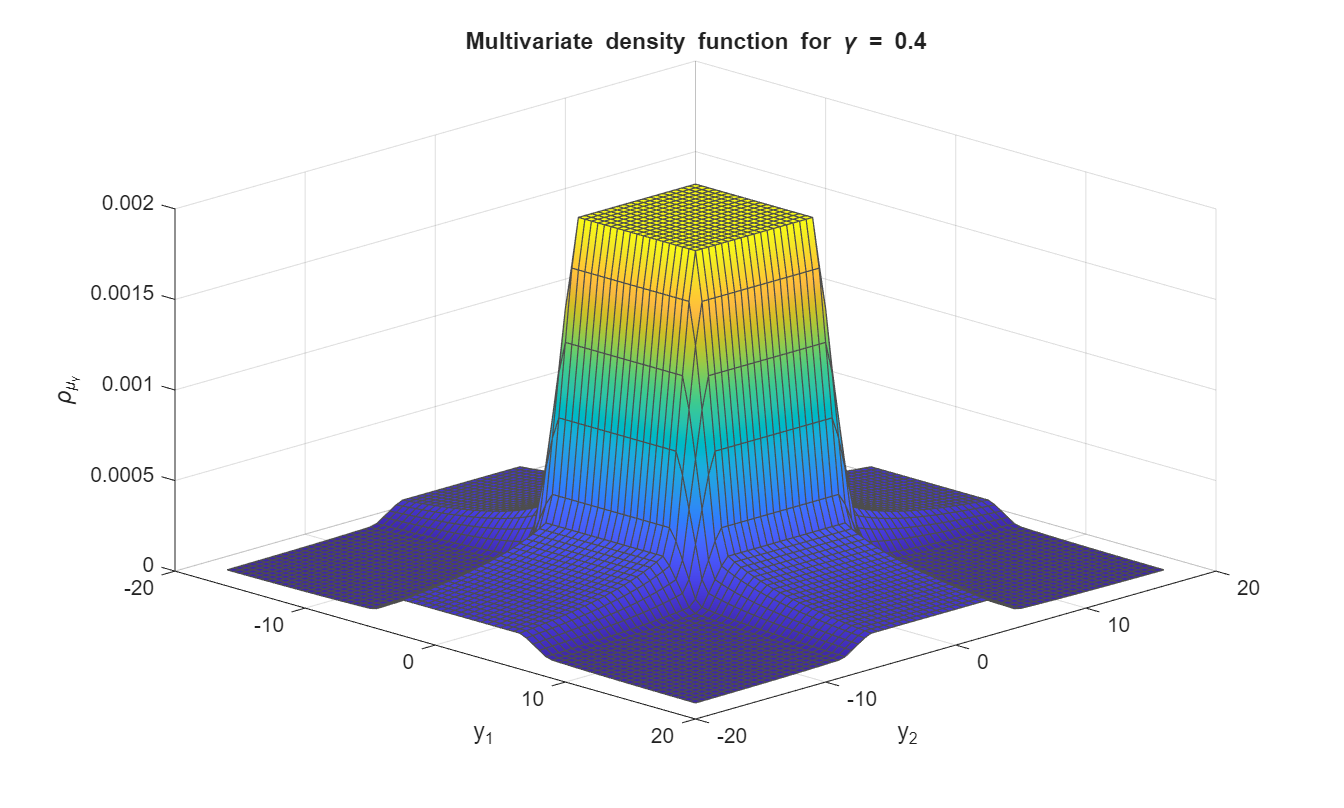}
\end{minipage}
\hfill
\begin{minipage}{0.48\textwidth}
\centering
\includegraphics[width=\textwidth]{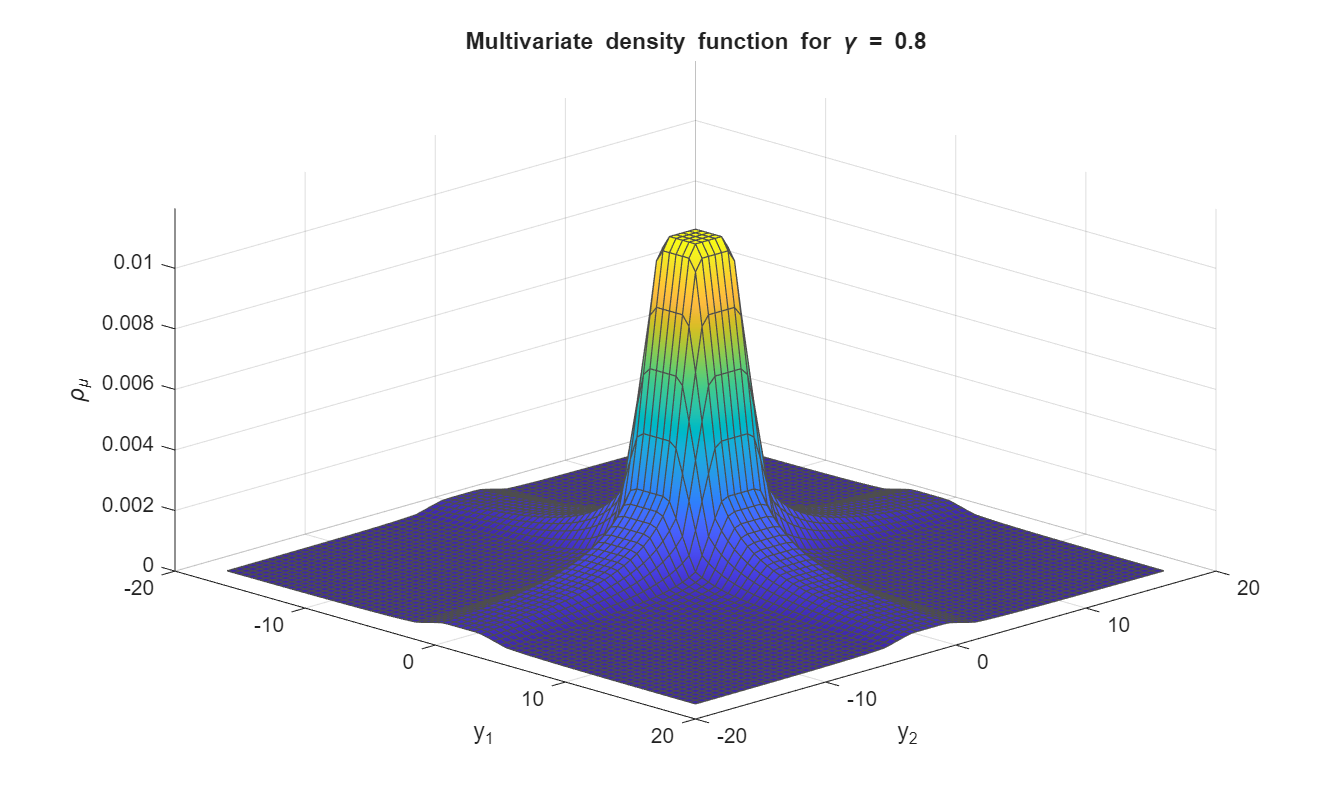} 
\end{minipage}
\caption{Plots of $\rho_{\mu_\gamma}$ (left) for  $\gamma=0.4$ and $\rho_{\mu_\gamma}$ (right) for $\gamma=0.8$.}
\label{fig:sigma_phi}
\end{figure}
\begin{figure}[H]
  \centering
  \includegraphics[width=0.9\textwidth]{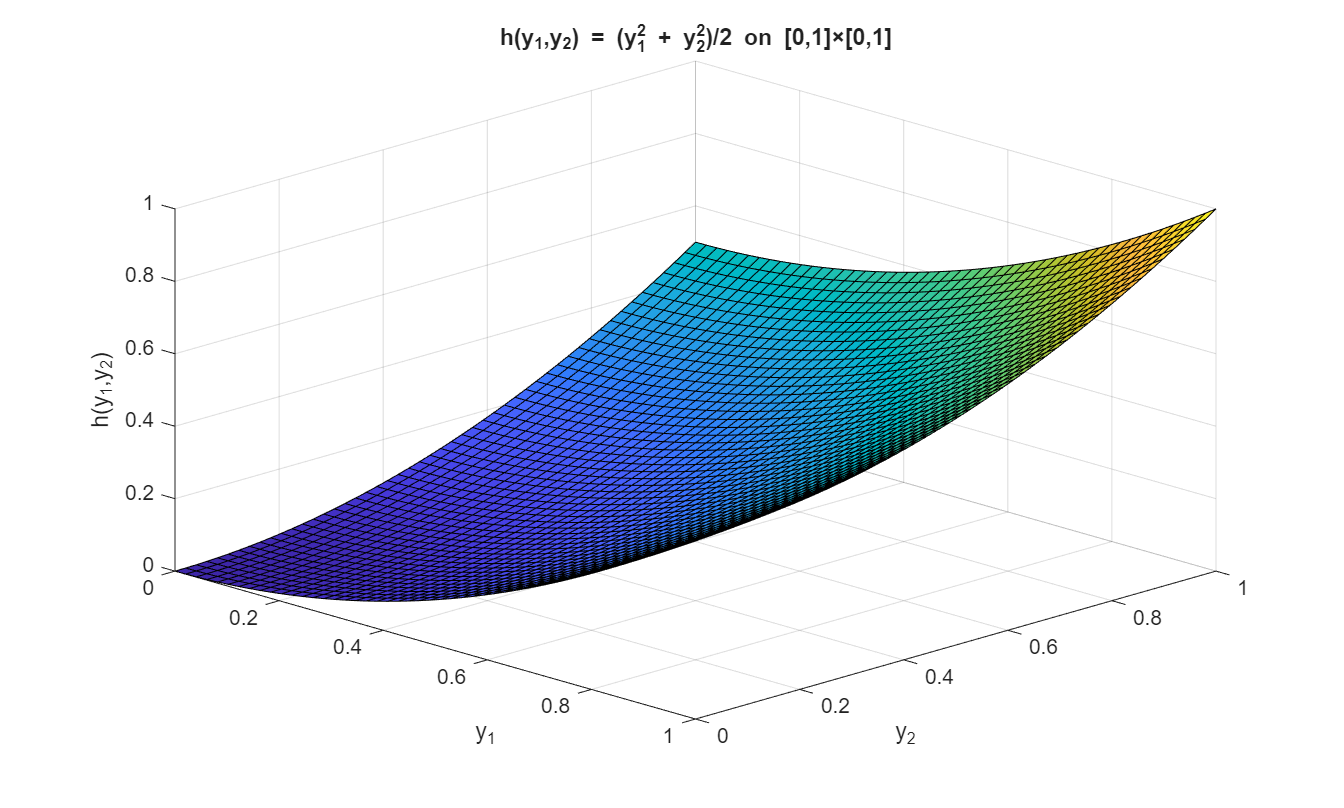}
  \caption*{Figure 4: Plot of the function h on $[0,1]^2$.}
\end{figure}
\begin{figure}[ht]
\centering
\begin{minipage}{0.48\textwidth}
\centering
\includegraphics[width=\textwidth]{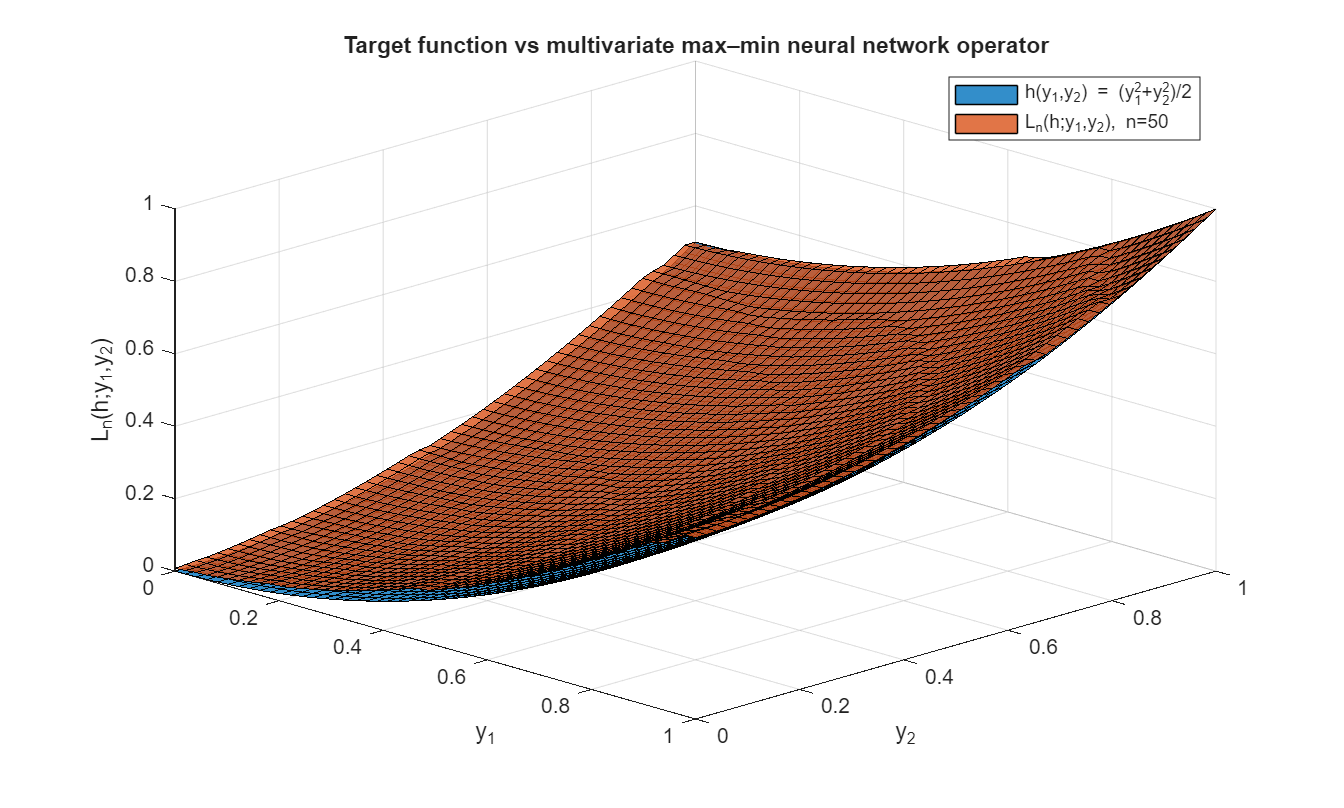}
\end{minipage}
\hfill
\begin{minipage}{0.48\textwidth}
\centering
\includegraphics[width=\textwidth]{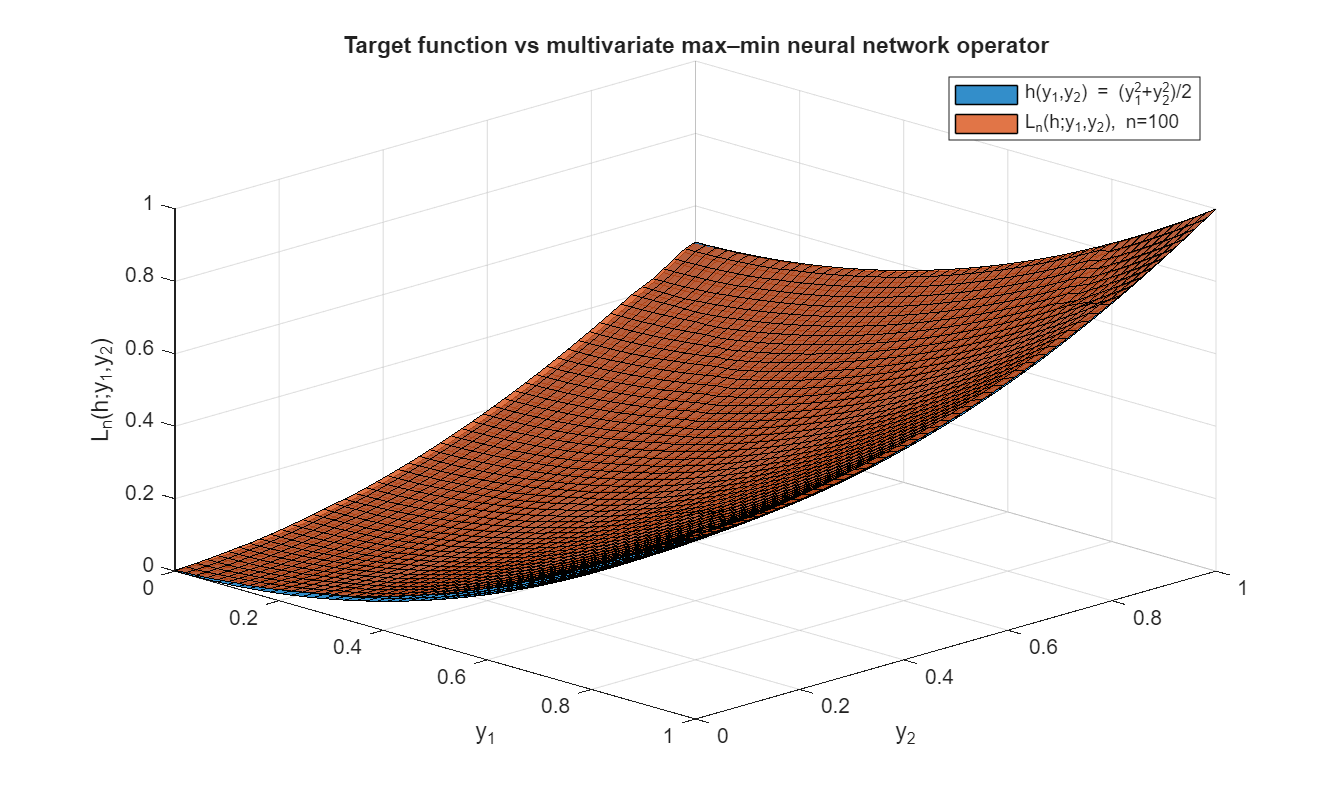} 
\end{minipage}
\caption*{Figure 5: Plots of $\mathcal{L}_n(h;\bar{y})$ activated by  logistic sigmoidal function (left) for $n=50$ and  (right)  for $n=100$.}
\label{fig:sigma_phi}
\end{figure}
\begin{figure}[ht]
\centering
\begin{minipage}{0.48\textwidth}
\centering
\includegraphics[width=\textwidth]{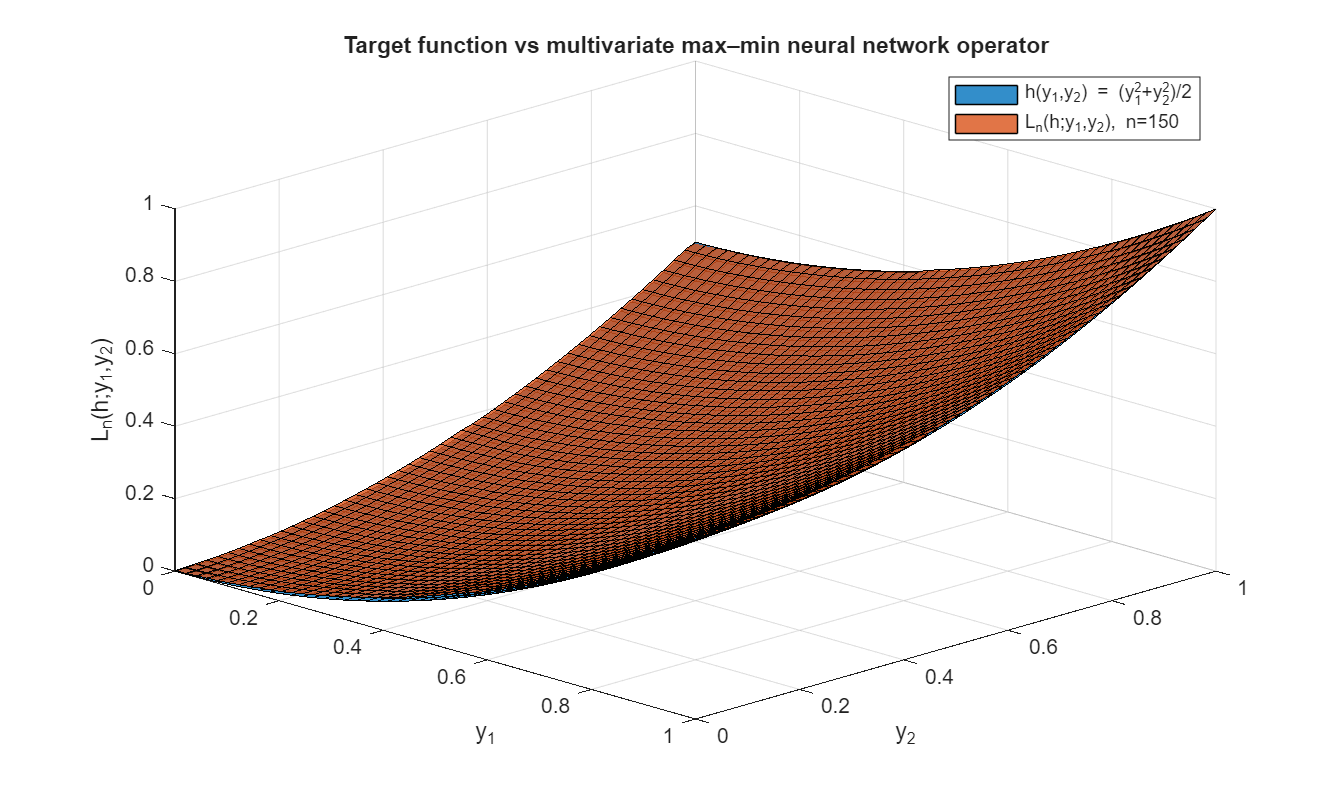}
\end{minipage}
\hfill
\begin{minipage}{0.48\textwidth}
\centering
\includegraphics[width=\textwidth]{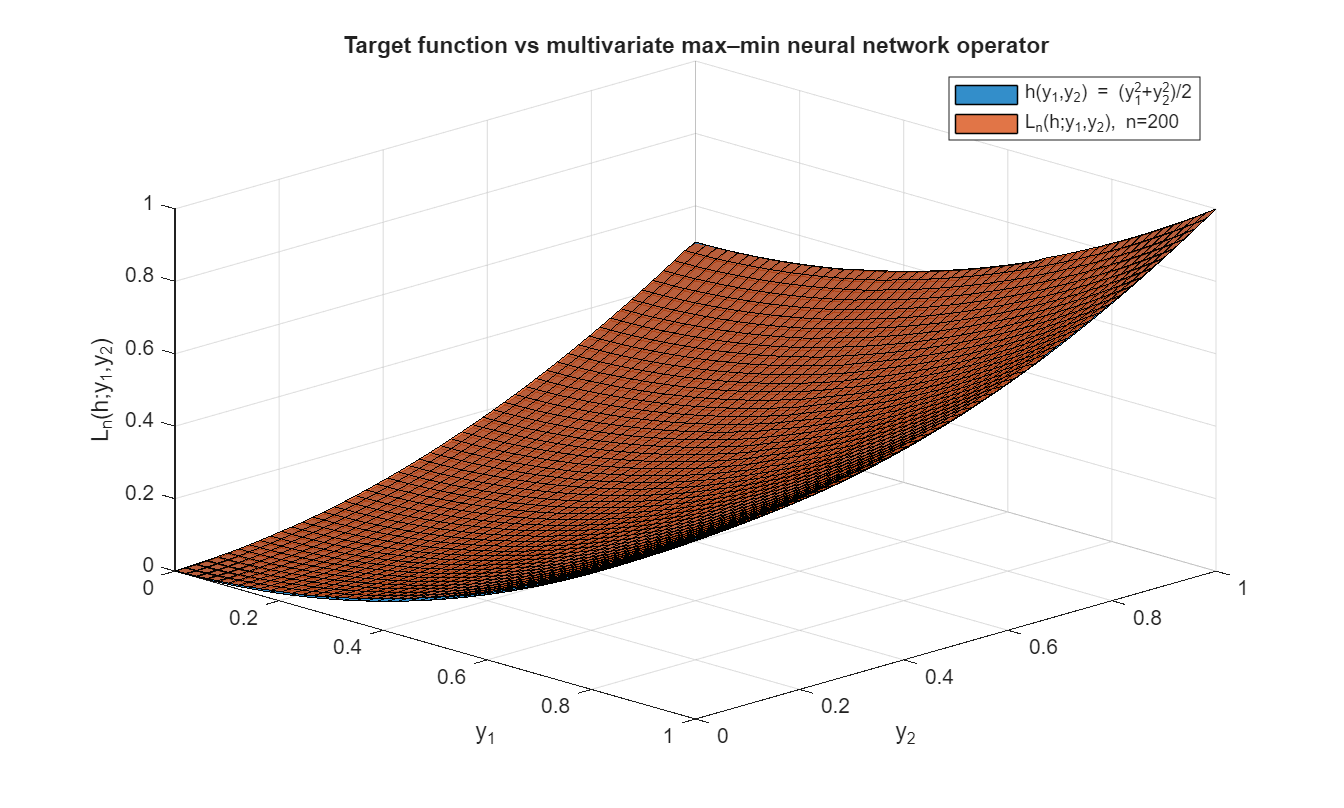} 
\end{minipage}
\caption*{Figure 6:  Plots of $\mathcal{L}_n(h;\bar{y})$ activated by  logistic sigmoidal function (left) for $n=150$ and  (right)  for $n=200$.}
\label{fig:sigma_phi}
\end{figure}
In Figure 2, we showed the plot of $\mu_\gamma$ and $\varphi_{\mu_\gamma}$ for
$\gamma = 0.4, 0.8$. Moreover, in Figure 3, we plotted the corresponding
multi-dimensional $\rho_{\mu_\gamma}$ for $r=2$.\\
Finally, we have 
\[
\sigma_{\alpha,\beta}(y) := e^{-\alpha e^{-\beta y}}, \qquad y \in \mathbb{R},
\]
where $\alpha>0$ and $\beta>0$, is the class of Gompertz
functions. Which are the sigmoidal function but not satisfying condition (a).\\
Now, in order to conclude this section and to illustrate the
performance of the above approximation processes, we give a
numerical example of a concrete approximation that can be
achieved by the above operators. We consider the following
continuous function on $[0,1]^2 $(see Figure\ 4),
$h(y_1,y_2) := \frac{y_1^2+y_2^2}{2}.$\\
In Figures 5 and 6, we consider approximations for $h$ by the multivariate max-min neural network operators $\mathcal{L}_n(h;\bar{y})$
activated by the logistic function for different values of $n$. We can observe from the figures that as the value of $n$ increases, the approximation becomes better.
\begin{center} 
\large{\textbf{Conclusion}} 
\end{center}
In this paper, we introduce well-defined multivariate max-min neural network operators together with their associated quasi-interpolation schemes. We analyse both pointwise and uniform convergence and examine the corresponding rates of convergence. The proposed multivariate max-min neural network operators model neuroprocessing systems in which the global network behavior is primarily governed by a single artificial neuron. These operators yield a constructive nonlinear approximation framework based on a class of neural networks that provides more accurate approximations than the linear counterparts studied in classical neural network operators.\\
More specifically, although the theoretical order of approximation coincides with that of classical neural network operators, multivariate max-min neural network operators exhibit superior global approximation quality. This improvement arises from sharper constants in the associated error estimates compared to those of classical operators. Finally, we present some examples of sigmoidal functions that illustrate the underlying assumptions and main results along with their graphical representation. We also discuss an example to show that the operator $\mathcal{L}_n(h;\bar{y})$ activated by the logistic function for different values of $n$ coincides with $h$ as $n$ increases.

\end{document}